\documentclass{article}
\usepackage{spconf}

\usepackage[latin1]{inputenc} 
\usepackage[pdftex]{graphicx}
\usepackage{amsmath,epsfig}
\usepackage{amssymb,cases}
\usepackage{subfigure}
\usepackage{tabularx,colortbl}
\usepackage[T1]{fontenc}
\usepackage{etex}
\usepackage[all,knot,arc,frame,matrix,arrow,graph,curve,color]{xy}
\usepackage{amssymb,amsthm}
\usepackage{leftidx}
\usepackage{graphics}
\usepackage{booktabs}
\usepackage{multicol}
\usepackage{multirow}
\usepackage{tkz-graph}
\usepackage{xcolor}
\usepackage{tikz}
\usepackage[linesnumbered, ruled, vlined]{algorithm2e}
\usepackage{algorithmic}
\usepackage{listings}

\usepackage{verbatim}

\usetikzlibrary{decorations.pathmorphing} 
\usetikzlibrary{fit}					
\usetikzlibrary{backgrounds}	
\usetikzlibrary{shadows}
\usetikzlibrary{mindmap}

\usetikzlibrary{positioning,shapes,shadows,arrows}

\usetikzlibrary{arrows,shapes}

\newcounter{definicao}
\renewcommand \thedefinicao
     {\ifnum \c@section>\z@ \fi \@arabic\c@definicao}
\def\fps@definicao{tbd}
\def\ftype@definicao{2}
\def\ext@definicao{lod}
\def\fnum@definicao{\definicaoname~\thedefinicao}

\newcounter{contProgram}

\newenvironment{definicao*}
               {\@dblfloat{definicao}}
               {\end@dblfloat}

\newcommand\definicaoname{{\bf Definition}}

\newcounter{teorema}

\renewcommand \theteorema
     {\ifnum \c@section>\z@ \fi \@arabic\c@teorema}
\def\fps@teorema{tbp}
\def\ftype@teorema{3}
\def\ext@teorema{lot}
\def\fnum@teorema{\teoremaname~\theteorema}

\newcounter{contTeorema}

\newenvironment{teorema*}
               {\@dblfloat{teorema}}
               {\end@dblfloat}

\newcommand\teoremaname{{\bf Theorem}}

\newcounter{propriedade}
\renewcommand \thepropriedade
     {\ifnum \c@section>\z@ \fi \@arabic\c@propriedade}
\def\fps@propriedade{tbp}
\def\ftype@propriedade{4}
\def\ext@propriedade{lop}
\def\fnum@propriedade{\propriedadename~\thepropriedade}

\newcounter{contpropriedade}

\newenvironment{propriedade*}
               {\@dblfloat{propriedade}}
               {\end@dblfloat}

\newcommand\propriedadename{{\bf Property}}

\newcounter{sjproposition}
\renewcommand \thesjproposition
     {\ifnum \c@section>\z@ \fi \@arabic\c@sjproposition}
\def\fps@sjproposition{tbp}
\def\ftype@sjproposition{5}
\def\ext@sjproposition{lopr}
\def\fnum@sjproposition{\sjpropositionname~\thesjproposition}

\newcounter{contsjproposition}

\newenvironment{sjproposition*}
               {\@dblfloat{sjproposition}}
               {\end@dblfloat}

\newcommand\sjpropositionname{{\bf Proposition }}

\newcounter{sjproof}
\renewcommand \thesjproof
     {\ifnum \c@section>\z@ \fi \@arabic\c@sjproof}
\def\fps@sjproof{tbp}
\def\ftype@sjproof{3}
\def\ext@sjproof{lop}
\def\fnum@sjproof{\sjproofname~\thesjproof}

\newcounter{contsjproof}

\newenvironment{sjproof*}
               {\@dblfloat{sjproof}}
               {\end@dblfloat}

\newcommand\sjproofname{\textbf{{\textit {Proof}}}}

\newcommand\vertices{V}
\newcommand\edges{E}

\newcommand\G{G}
\newcommand\W{w}

\newcommand\SCALE[2]{S(#1,#2)}
\newcommand\SCALESET[1]{{{L}({#1})}}
\newcommand\SCALESETSYMBOL{{{L}}}

\newcommand\MST[1][\G]{{MST}}
\newcommand\edgeValue[1]{w(#1)}

\newcommand\path[2]{\Pi_{#1}^{#2}}

\newcommand{\ELIMINE}[1]{}
\newcommand{\ELIMINEPR}[1]{} 
\newcommand{\Vfunc}[1]{F^\star}

\newcommand{\Proof}[1]{}

\newcommand{\Nset}[0]{\mathbb{N}}

\newcommand{\Fset}[1]{\mathcal{F}}

\newcommand\CompX{X}
\newcommand\CompY{Y}
\newcommand\DIFF{\ensuremath{\mathit{Diff}}\xspace({\CompX},{\CompY})}

\newcommand\INT[1]{\ensuremath{\mathit{Int}}\xspace({#1})}
\newcommand\RINT[2][]{S_{#1}({#2})}
\newcommand\NRINT[2][]{{S'}_{#1}({#2})}

\newcommand\Lowlevel[1]{{L}(#1)}
\newcommand\COMMENT{}

\newcommand\PARTITIONSETGW[3][]{{\cal P}_{#1}^{#3}}

\newcommand\tree{T}
\newcommand\HIERARCHYW[1]{{\cal H}^{#1}}

\renewcommand\COMMENT[1]{}

\tikzstyle{background2}=[rectangle,fill=black!10!white,inner sep=0.2cm,rounded corners=1mm]
\tikzstyle{vertexNULL}=[rectangle,minimum size=20pt,inner sep=0pt]
\tikzstyle{vertex}=[circle,fill=black!25,minimum size=20pt,inner sep=0pt]

\tikzstyle{vertexG}=[circle,fill=black!40,text=white,minimum size=15pt,inner sep=0pt]
\tikzstyle{edgeG} = [draw,thick,-]
\tikzstyle{edgeMG} = [draw,dotted,thick,-]
\tikzstyle{edgeEMG} = [draw,dashed,thick,-]

\tikzstyle{order}=[rectangle,fill=blue!5,minimum size=20pt,inner sep=0pt]
\tikzstyle{vertexNAME}=[rectangle,minimum size=20pt,inner sep=0pt]

\tikzstyle{vertexCT}=[rectangle,fill=green!05,minimum size=20pt,inner sep=0pt]
\tikzstyle{vertexKCT}=[circle,fill=red!45,minimum size=15pt, inner sep=0pt]
\tikzstyle{selected vertex} = [vertex, fill=red!24]
\tikzstyle{partition 1} = [vertex, fill=red!24]
\tikzstyle{partition 2} = [vertex, fill=blue!24]
\tikzstyle{partition 3} = [vertex, fill=green!24]
\tikzstyle{partition 4} = [vertex, fill=gray!24]
\tikzstyle{partition 5} = [vertex, fill=red!70]
\tikzstyle{partition 6} = [vertex, fill=yellow!50]
\tikzstyle{partition 7} = [vertex, fill=yellow!20]

\tikzstyle{edgeCT} = [draw,thick,-,red!50]
\tikzstyle{edge} = [draw,line width=7pt,-,blue!50]
\tikzstyle{edge initial} = [draw,thick,-]
\tikzstyle{edge initial CT} = [draw,thick,->]
\tikzstyle{edge initial KCT} = [draw,thick,->,red!20]

\tikzstyle{weight} = [font=\small]
\tikzstyle{selected edge} = [draw,line width=5pt,-,red!50]
\tikzstyle{ignored edge} = [draw,line width=5pt,-,black!20]

\tikzstyle{abstractSJ}=[rectangle, draw=black, rounded corners, fill=blue!40, drop shadow,
        text centered, anchor=north, text=white, text width=0.5cm,minimum size=20pt]
\tikzstyle{commentSJ}=[anchor=center,text=white]
\tikzstyle{myarrowSJ}=[->, >=open triangle 90, thick]
\tikzstyle{lineSJ}=[-, thick]

\title{An efficient hierarchical graph based image segmentation}
\twoauthors
  {Silvio Jamil F. Guimarães$^{\sharp\S}$}
{$^\sharp$VIPLAB - ICEI - PUC Minas \\ 
sjamil{@}pucminas.br
} 
{
  Jean Cousty$^\S$, Yukiko Kenmochi$^\S$ 
  and Laurent Najman$^\S$
  }
	{$^\S$Université Paris-Est, LIGM, ESIEE - UPEMLV  - CNRS\\
	\{j.cousty,y.kenmochi,l.najman\} {@}esiee.fr
	}

\begin{document}
\ninept
\maketitle

\begin{abstract}
  Hierarchical image segmentation provides region-oriented
  scale-space, {\em i.e.}, a set of image segmentations at different
  detail levels in which the segmentations at finer levels are nested
  with respect to those at coarser levels. Most image segmentation
  algorithms, such as region merging algorithms, rely on a criterion
  for merging that does not lead to a hierarchy, and for which the
  tuning of the parameters can be difficult.  In this work, we propose
  a hierarchical graph based image segmentation relying on a
  criterion popularized by Felzenzwalb and Huttenlocher.  We illustrate
  with both real and synthetic images, showing efficiency, ease of
  use, and robustness of our method.
\end{abstract}

\vspace{-0.5cm}

\section{Introduction} \label{sec:intro}

Image segmentation is the process of grouping perceptually similar
pixels into regions. A hierarchical image segmentation is a set of
image segmentations at different detail levels in which the
segmentations at coarser detail levels can be produced from simple
merges of regions from segmentations at finer detail
levels. Therefore, the segmentations at finer levels are nested with
respect to those at coarser levels. Hierarchical methods have the
interesting property of preserving spatial and neighboring information
among segmented regions. Here, we propose a hierarchical image
segmentation in the framework of edge-weighted graphs, where the image
is equipped with an adjacency graph and the cost of an edge is given
by a dissimilarity between two points of the image.

\COMMENT{\textbf{\cite{JS71,Gower1969} stating an equivalence between
    hierarchies and minimum spanning trees (MST), however the first
    appearance for classification in pattern recognition data from the
    seminal work of Zahn \cite{Zahn1971}. Its use for image
    segmentation was introduced by Morris {\em et al.}
    \cite{Morris1986} in 1986 and popularized in 2004 by Felzenswalb
    and Huttenlocher \cite{Felzenszwalb2004}.  In \cite{Cousty2011},
    it was studied some optimality properties of hierarchical
    segmentations. }}

Any hierarchy can be represented with a minimum spanning tree. The
first appearance of this tree in pattern recognition dates back to the
seminal work of Zahn \cite{Zahn1971}. Lately, its use for image
segmentation was introduced by Morris {\em et al.}  \cite{Morris1986}
in 1986 and popularized in 2004 by Felzenswalb and Huttenlocher
\cite{Felzenszwalb2004}. However the region-merging method
\cite{Felzenszwalb2004} does not provide a hierarchy.  In
\cite{Najman2011,Cousty2011}, it was studied some optimality
properties of hierarchical segmentations. Considering that, for a
given image, one can tune the paramaters of the well-known method
\cite{Felzenszwalb2004} for obtaining a correct segmentation of this
image. We provide in this paper a hierachical version of this method
that removes the need for parameter tuning. 

The algorithm of \cite{Felzenszwalb2004} is the following. First, a
minimum spanning tree (MST) is computed, and all the decisions are
taken on this tree. For each edge linking two vertices $x$ and $y$,
following a non-decreasing order of their weights, the following steps
are performed:
\begin{itemize} \addtolength{\itemsep}{-0.2\baselineskip}
\item [(i)] Find the region $\CompX$ that contains $x$.
\item [(ii)] Find the region $\CompY$ that contains $y$.
\item [(iii)] Merge $\CompX$ and $\CompY$ according to a certain
  criterion. 
\end{itemize}

The criterion for region-merging in \cite{Felzenszwalb2004} measures
the evidence for a boundary between two regions by comparing two
quantities: one based on intensity differences across the boundary,
and the other based on intensity differences between neighboring pixel
within each region. More precisely, in step (iii), in order to know
whether two regions must be merged, two measures are considered. The
\textit{internal difference} $\INT{\CompX}$ of a region $\CompX$ is
the highest weight of an edge linking two vertices of $\CompX$ in the
MST.  The \textit{difference} $\DIFF$ between two neighboring regions
$\CompX$ and $\CompY$ is the smallest weight of an edge that links
$\CompX$ to $\CompY$. Then, two regions $\CompX$ and $\CompY$ are
merged when:
\begin{equation} \footnotesize
\DIFF\leq\min\{\INT{\CompX}+\frac{k}{|\CompX|},\INT{\CompY}+\frac{k}{|\CompY|}\} \label{eq:diff}
\end{equation} 
\noindent where $k$ is a parameter allowing to prevent the merging of
large regions (\textit{i.e.}, larger $k$ force smaller regions to be
merged).

The merging criterion defined by Eq.~\eqref{eq:diff} depends on the
scale $k$ at which the regions $\CompX$ and $\CompY$ are
observed. More precisely, let us consider the \textit{(observation)
  scale $\RINT[\CompY]{\CompX}$ of $\CompX$ relative to $\CompY$} as a
measure based on the difference between $\CompX$ and $\CompY$, on the
internal difference of $\CompX$ and on the size |$\CompX$| of
$\CompX$:
\begin{eqnarray} \footnotesize
\RINT[\CompY]{\CompX}=(\DIFF-\INT{\CompX})\times~|\CompX|.\label{eq:scale}
\end{eqnarray}
\vspace{-0.5cm}

\noindent Then, the scale $\SCALE{\CompX}{\CompY}$ is simply defined as:
\begin{eqnarray} \footnotesize
\SCALE{\CompX}{\CompY}=\max (\RINT[\CompY]{\CompX},\RINT[\CompX]{\CompY}).
\label{eq:diss}
\end{eqnarray}
Thanks to this notion of a scale Eq.~\eqref{eq:diff} can be written as:
\begin{eqnarray} \footnotesize
k\geq\SCALE{\CompX}{\CompY}.
\label{eq:diss:knew}
\end{eqnarray}
\noindent In other words, Eq.\eqref{eq:diss:knew} states that the
neighboring regions $\CompX$ and $\CompY$ merge when their scale is
less than the threshold parameter $k$. 

\begin{figure}
\begin{center}
\subfigure[Original]{
\includegraphics[width=0.30\linewidth]{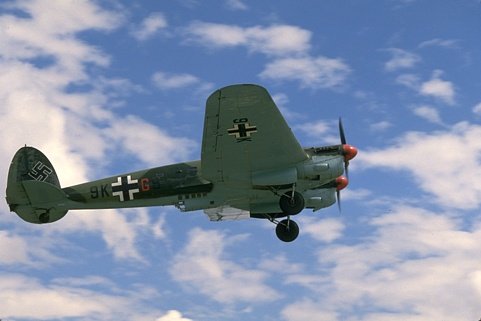}
}
\subfigure[$k=7500~(8)$]{
\includegraphics[width=0.30\linewidth]{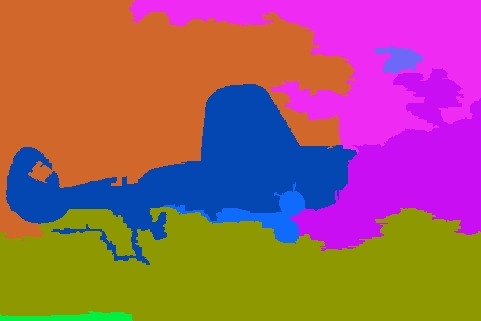}
}
\subfigure[$k=9000~(14)$]{
\includegraphics[width=0.30\linewidth]{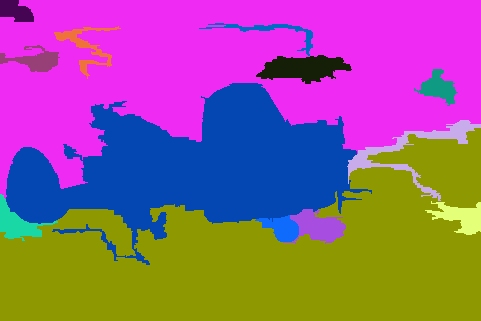}
}
\end{center}
\caption{A real example illustrating the violation of the causality
  principle by \cite{Felzenszwalb2004}: the number of regions (in
  parentheses) increases from 8 to 14, instead of decreasing when the
  so-called ``scale of observation'' increases. 
}
\label{pedro:real:nonhierarchical}

\end{figure}

Even if the image segmentation results obtained by the method proposed
in \cite{Felzenszwalb2004} are interesting, the user faces two major
issues:
\begin{itemize}
\item first, the number of regions may increase when the parameter $k$
  increases. This should not be possible if $k$ was a true scale of
  observation: indeed, it violates the {\em causality principle} of
  multi-scale analysis, that states in our case \cite{Guigues2006}
  that a contour present at a scale $k_1$ should be present at any
  scale $k_2<k_1$. Such a behaviour is demonstrated on
  Fig.~\ref{pedro:real:nonhierarchical}.
\item Second, even when the number of regions decreases, contours
  are not stable: they can move when the parameter $k$ varies,
  violating a {\em location principle}. Such a situation is
  illustrated on Fig.~\ref{pedro:nonhierarchical}.
\end{itemize}
Given these two issues, the tunning of the parameters of
\cite{Felzenszwalb2004} is a difficult task.

Following \cite{Guigues2006}, we believe that, in order for $k$ to be
a true scale-parameter, we have to satisfy both the causality
principle and the location principle, which leads to work with
hierarchy of segmentations. Reference \cite{Haxhimusa2004} is the first to
propose an algorithm producing a hierarchy of segmentations based on
\cite{Felzenszwalb2004}. However, this method is an iterative version
of \cite{Felzenszwalb2004} that uses a threshold function, and
requires a tunning of the threshold parameter.


\tikzstyle{vertexNULL}=[rectangle,minimum size=20pt,inner sep=0pt]
\tikzstyle{vertex}=[circle,fill=black!25,draw=black,minimum size=20pt,inner sep=0pt]
\tikzstyle{order}=[rectangle,fill=blue!5,minimum size=20pt,inner sep=0pt]

\tikzstyle{vertexRegionA}=[circle,fill=black!10,draw=black,minimum size=20pt,inner sep=0pt]
\tikzstyle{vertexRegionB}=[circle,fill=white!100,draw=black,minimum size=20pt,inner sep=0pt]
\tikzstyle{vertexRegionC}=[circle,fill=black!70,text=white,draw=black,minimum size=20pt,inner sep=0pt]

\tikzstyle{vertexCT}=[rectangle,fill=green!05,minimum size=20pt,inner sep=0pt]
\tikzstyle{vertexKCT}=[circle,fill=red!45,minimum size=15pt, inner sep=0pt]
\tikzstyle{selected vertex} = [vertex, fill=red!24]
\tikzstyle{partition 1} = [vertex, fill=red!24]
\tikzstyle{partition 2} = [vertex, fill=blue!24]
\tikzstyle{partition 3} = [vertex, fill=green!24]
\tikzstyle{partition 4} = [vertex, fill=gray!24]
\tikzstyle{partition 5} = [vertex, fill=red!70]
\tikzstyle{partition 6} = [vertex, fill=yellow!50]
\tikzstyle{partition 7} = [vertex, fill=yellow!20]

\tikzstyle{edgeCT} = [draw,thick,-,red!50]
\tikzstyle{edge} = [draw,line width=7pt,-,blue!50]
\tikzstyle{edge initial} = [draw,thick,-]
\tikzstyle{edge initial CT} = [draw,thick,->]
\tikzstyle{edge initial KCT} = [draw,thick,->,red!20]

\tikzstyle{weight} = [font=\small]
\tikzstyle{selected edge} = [draw,line width=5pt,-,red!50]
\tikzstyle{ignored edge} = [draw,line width=5pt,-,black!20]

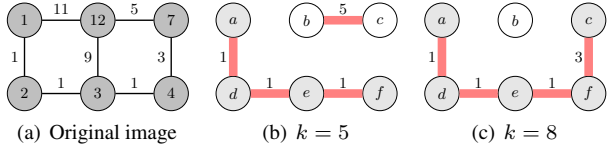
\begin{figure}[here]
\begin{center}
\subfigure[Original image]{
\scalebox{0.65}[0.65]{
\begin{tikzpicture}[scale=1.5, auto,swap]
    \foreach \pos/\name /\pixel in {{(0,0)/a/1}, {(1,0)/b/12}, {(2,0)/c/7},
                            {(0,-1)/d/2}, {(1,-1)/e/3}, {(2,-1)/f/4}}
        \node[vertex] (\name) at \pos {$\pixel$};
        
    \foreach \source/ \dest /\weight  in {a/d/1, b/a/11,
    																		  c/b/5,b/e/9,
                                          c/f/3, f/e/1,
                                          e/d/1}
        \path[edge initial] (\source) -- node[weight] {$\weight$} (\dest);
\end{tikzpicture}
}
}
\subfigure[$k=5$]{
\scalebox{0.65}[0.65]{
\begin{tikzpicture}[scale=1.5, auto,swap]
    \foreach \pos/\name in {{(0,0)/a},
                            {(0,-1)/d}, {(1,-1)/e}, {(2,-1)/f}}
        \node[vertexRegionA] (\name) at \pos {$\name$};
        
    \foreach \pos/\name in {{(1,0)/b}, {(2,0)/c}}
        \node[vertexRegionB] (\name) at \pos {$\name$};

    \foreach \source/ \dest /\weight  in {a/d/1, 
    																		  c/b/5,
                                          f/e/1,
                                          e/d/1}
        \path[edge initial] (\source) -- node[weight] {$\weight$} (\dest);
    \foreach \source/ \dest /\weight  in {a/d/1, 
    																		  c/b/5,
                                          f/e/1,
                                          e/d/1}
        \path[selected edge] (\source) -- node[weight] {} (\dest);
\end{tikzpicture}
}
}
\subfigure[$k=8$]{
\scalebox{0.65}[0.65]{
\begin{tikzpicture}[scale=1.5, auto,swap]
    \foreach \pos/\name in {{(2,0)/c}, {(0,0)/a},
                            {(0,-1)/d}, {(1,-1)/e}, {(2,-1)/f}}
        \node[vertexRegionA] (\name) at \pos {$\name$};
        
    \foreach \pos/\name in {{(1,0)/b}}
        \node[vertexRegionB] (\name) at \pos {$\name$};
    \foreach \source/ \dest /\weight  in {a/d/1, 
                                          c/f/3, 
                                          f/e/1,
                                          e/d/1}
        \path[edge initial] (\source) -- node[weight] {$\weight$} (\dest);

    \foreach \source/ \dest /\weight  in {a/d/1, 
                                          c/f/3, 
                                          f/e/1,
                                          e/d/1}
        \path[selected edge] (\source) -- node[weight] {} (\dest);
\end{tikzpicture}
}
}

%
\end{center}

\vspace{-0.5cm}
\caption{\label{pedro:nonhierarchical}An example illustrating the
  violation of the location property by \cite{Felzenszwalb2004}: the
  contours are unstable from one ``scale'' to another. }
\end{figure}

The main result of this paper is an efficient hierarchical image
segmentation algorithm based on the dissimilarity measure of
\cite{Felzenszwalb2004}. Our algorithm has a computational cost
similar to \cite{Felzenszwalb2004}, but provides all scales of
observations instead of only one segmentation level. As it is a
hierarchy, the result of our algorithm satisfies both the locality
principle and the causality principle. In particular, and in contrast
with \cite{Felzenszwalb2004}, the number of regions is decreasing when
the scale parameter increases, and the contours do not move from one
scale to another. 

\begin{figure} [here]
\begin{center}
\subfigure[Saliency map]{
\includegraphics[width=0.30\linewidth]{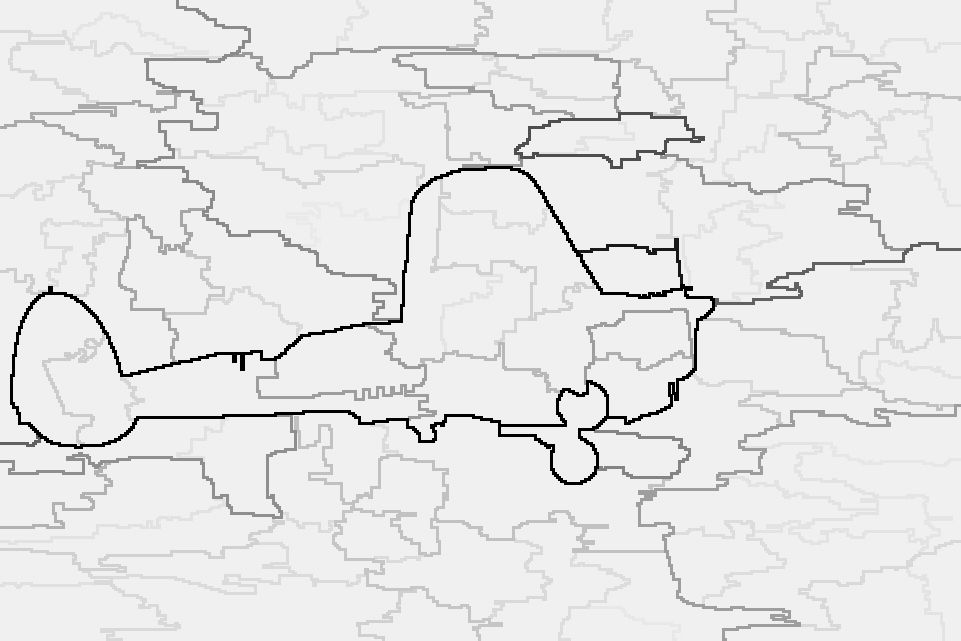}
}
\subfigure[$k=1000~(22)$]{
\includegraphics[width=0.30\linewidth]{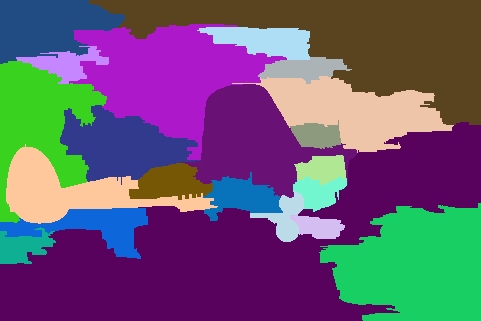}
}
\subfigure[$k=5000~(6)$]{
\includegraphics[width=0.30\linewidth]{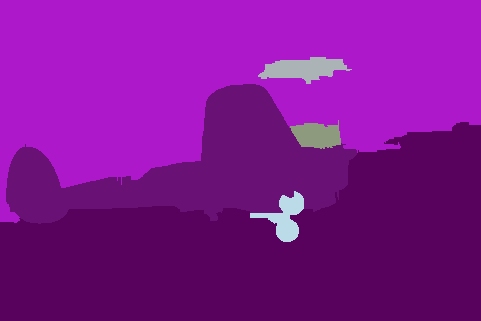}
}
\end{center}
.
\caption{A real example illustrating the saliency map of
  Fig.~\ref{pedro:real:nonhierarchical}(a) computed with our
  approach. We display in (b) and (c) two image segmentations extracted from the
  hierarchy at scales $1000$ and $5000$, together with their number of
  regions (in parentheses).\label{our:real:hierarchical}}
\end{figure}

Figure~\ref{our:real:hierarchical} illustrates  the results obtained by applying our method to the same image of Fig.~\ref{pedro:real:nonhierarchical}(a), with
segmentations at two different scales of observations, as well as a
saliency map \cite{Najman1996,Najman2011,Cousty2011} (a map indicating
the disparition level of contours and whose thresholds give the set of
all segmentations).


This work is organized as follows. In Section~\ref{sec:algo}, we
present our hierarchical method for color image segmentation. Some
experimental results are given in Section~\ref{sec:experiment}.
Finally, in Section~\ref{sec:conclusions}, some conclusions are drawn
and further works are discussed.

\section{An efficient hierarchical graph based image segmentation}
\label{sec:algo}

In this section, we describe our method to compute a hierarchy of
partitions based on observation scales as defined by
Eq.~\ref{eq:diss}.  Let us first recall some important notions for
handling hierarchies \cite{Morris1986,Najman2011,Cousty2011}.

To every tree $\tree$ spanning the set $\vertices$ of the image
pixels, to every map $\W:\edges\rightarrow\Nset$ that weights the
edges of $\tree$ and to every threshold $\lambda \in \Nset$, one may
associate the partition $\PARTITIONSETGW[\lambda]{\G}{\W}$ of
$\vertices$ induced by the connected components of the graph made by
$\vertices$ and the edges of weight below $\lambda$.  \COMMENT{We
  denote by $\PARTITIONSETGW[\lambda]{\G}{\W}(x)$ be an element of the
  partition containing the vertex $x$.}  It is well known
\cite{Morris1986,Cousty2011} that for any two values $\lambda_1$ and
$\lambda_2$ such that $\lambda_1\geq\lambda_2$, the partitions
$\PARTITIONSETGW[\lambda_1]{\G}{\W}$ and
$\PARTITIONSETGW[\lambda_2]{\G}{\W}$ are \textit{nested} and
$\PARTITIONSETGW[\lambda_1]{\G}{\W}$ is \textit{coarser} than
$\PARTITIONSETGW[\lambda_2]{\G}{\W}$. Hence, the set
$\HIERARCHYW{\W}=\langle\PARTITIONSETGW[\lambda]{\G}{\W}~|~\lambda\in\Nset\rangle$
is a {\em hierarchy of partitions induced by the weight map $\W$}.

Our algorithm does not explicitly produce a hierarchy of partitions,
but instead it produces a weight map $\SCALESETSYMBOL$ (scales of
observations) from which the desired hierarchy
$\HIERARCHYW{\SCALESETSYMBOL}$ can be infered. It starts from a
minimum spanning tree $\tree$ of the edge-weighted graph built from
the image. In order to compute the scale $\SCALESET{e}$ associated
with each edge of $\tree$, our method iteratively considers the edges
of $\tree$ in a non-decreasing order of their weights. For every edge
$e$, the weight map $\SCALESET{e}$ is initialized to $\infty$; then
for each edge $e$ linking two vertices $x$ and $y$ the following steps
are performed:

\begin{itemize}  
\item [(i)] Find the the region $\CompX$ of $\PARTITIONSETGW[\edgeValue{e}]{\G}{\W}$ that contains $x$.
\item [(ii)] Find the the region $\CompY$ of $\PARTITIONSETGW[\edgeValue{e}]{\G}{\W}$ that contains $y$.
\item [(iii)] Compute the hierarchical observation scale $\Lowlevel{e}$.
\end{itemize}

\noindent At step (iii), the \textit{hierarchical scale}
$\NRINT[\CompY]{\CompX}$ of $\CompX$ relative to $\CompY$ is needed to
obtain the value$~\Lowlevel{e}$. Intuitively, $\NRINT[\CompY]{\CompX}$
is the lowest observation scale at which some sub-region of $\CompX$,
namely $\CompX^{*}$, will be merged to $\CompY$. More precisely, using
an internal parameter $v$, this scale is computed as follows:
\begin{itemize}  
\item [(1)] Initialize the value of $v$ to 0.
\item [(2)] Increment the value of $v$ by 1.
\item [(3)] Find the the region $\CompX^{*}$ of $\PARTITIONSETGW[v]{\G}{\SCALESETSYMBOL}$ that contains $x$.
\item [(4)] Repeat steps 2 and 3 while $\RINT[\CompY]{\CompX^{*}}>~v$
\item [(5)] $\NRINT[\CompY]{\CompX}=v$.
\end{itemize}

\noindent With the appropriate changes, the same algorithm allows
$\NRINT[\CompX]{\CompY}$ to be computed. Then, the hierarchical scale
$\SCALESETSYMBOL(e)$ is simply set to:
\begin{equation}
\SCALESETSYMBOL(e) = \max\{\NRINT[\CompY]{\CompX},\NRINT[\CompX]{\CompY}\}.\label{eq:new:scale}
\end{equation}

\COMMENT{Comparing Eqs.~\ref{eq:diss} and ~\ref{eq:new:scale},
$\SCALESETSYMBOL(e)$ can be considered as a new scale to observe
diffence between $\CompX$ and $\CompY$.}

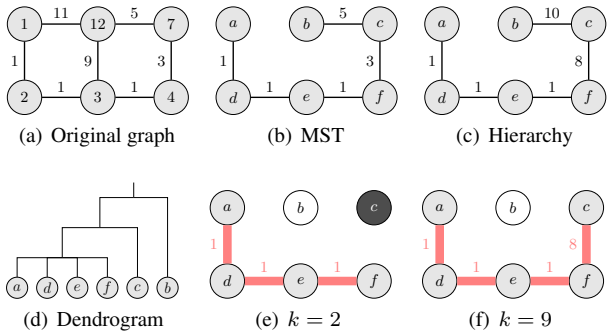
\begin{figure}
\begin{center}
\subfigure[Original graph]{
\scalebox{0.65}[0.65]{
\begin{tikzpicture}[scale=1.5, auto,swap]
    \foreach \pos/\name /\pixel in {{(0,0)/a/1}, {(1,0)/b/12}, {(2,0)/c/7},
                            {(0,-1)/d/2}, {(1,-1)/e/3}, {(2,-1)/f/4}}
        \node[vertexRegionA] (\name) at \pos {$\pixel$};
        
    \foreach \source/ \dest /\weight  in {a/d/1, b/a/11,
    																		  c/b/5,b/e/9,
                                          c/f/3, f/e/1,
                                          e/d/1}
        \path[edge initial] (\source) -- node[weight] {$\weight$} (\dest);
\end{tikzpicture}
}
}
\subfigure[MST]{
\scalebox{0.65}[0.65]{
\begin{tikzpicture}[scale=1.5, auto,swap]
    \foreach \pos/\name in {{(0,0)/a}, {(1,0)/b}, {(2,0)/c},
                            {(0,-1)/d}, {(1,-1)/e}, {(2,-1)/f}}
        \node[vertexRegionA] (\name) at \pos {$\name$};
    \foreach \source/ \dest /\weight  in {a/d/1, 
    																		  c/b/5,
                                          c/f/3, f/e/1,
                                          e/d/1}
        \path[edge initial] (\source) -- node[weight] {$\weight$} (\dest);
\end{tikzpicture}
}
}
\subfigure[Hierarchy]{
\scalebox{0.65}[0.65]{
\begin{tikzpicture}[scale=1.5, auto,swap]
    \foreach \pos/\name in {{(0,0)/a}, {(1,0)/b}, {(2,0)/c},
                            {(0,-1)/d}, {(1,-1)/e}, {(2,-1)/f}}
        \node[vertexRegionA] (\name) at \pos {$\name$};
    \foreach \source/ \dest /\weight  in {a/d/1, 
    																		  c/b/10,
                                          c/f/8, f/e/1,
                                          e/d/1}
        \path[edge initial] (\source) -- node[weight] {$\weight$} (\dest);
\end{tikzpicture}
}
}
\subfigure[Dendrogram]{
\scalebox{0.4}[0.4]{
\begin{tikzpicture}[scale=0.5]

    \node (a) [vertexRegionA] at (0,0)
        {
            {\Large $a$}
        };
    \node (d) [vertexRegionA] at (2,0)
        {
            {\Large $d$}
        };
    \node (e) [vertexRegionA] at (4,0)
        {
            {\Large $e$}
        };
    \node (f) [vertexRegionA] at (6,0)
        {
            {\Large $f$}
        };
    \node (c) [vertexRegionA] at (8,0)
        {
 						{\Large $c$}
        };
    \node (b) [vertexRegionA] at (10,0)
        {
            {\Large $b$}
        };


    \node (adef) [commentSJ] at (3,2)
        {
            \textbf{AB}
            \nodepart{second}name
        };
    \node (adefc) [commentSJ] at (5.5,4)
        {
            \textbf{AB}
            \nodepart{second}name
        };
    \node (adefcb) [commentSJ] at (7.75,6)
        {
            \textbf{AB}
            \nodepart{second}name
        };
    \node (final) [commentSJ] at (7.75,7)
        {
            \textbf{AB}
            \nodepart{second}name
        };

    \draw[lineSJ] (a.north)  |- (adef.center);
    \draw[lineSJ] (d.north)  |- (adef.center);
    \draw[lineSJ] (e.north)  |- (adef.center);
    \draw[lineSJ] (f.north)  |- (adef.center);
    \draw[lineSJ] (adef.center)  |- (adefc.center);
    \draw[lineSJ] (c.north)  |- (adefc.center);
    \draw[lineSJ] (b.north)  |- (adefcb.center);
    \draw[lineSJ] (adefc.center)  |- (adefcb.center);
    \draw[lineSJ] (adefcb.center)  |- (final.center);


\end{tikzpicture}
}
}
\subfigure[$k=2$]{
\scalebox{0.65}[0.65]{
\begin{tikzpicture}[scale=1.5, auto,swap]
    \foreach \pos/\name in {{(0,0)/a},
                            {(0,-1)/d}, {(1,-1)/e}, {(2,-1)/f}}
        \node[vertexRegionA] (\name) at \pos {$\name$};
    \foreach \pos/\name in {{(1,0)/b}}
        \node[vertexRegionB] (\name) at \pos {$\name$};
    \foreach \pos/\name in {{(2,0)/c}}
        \node[vertexRegionC] (\name) at \pos {$\name$};

    \foreach \source/ \dest /\weight  in {a/d/1, 
                                          f/e/1,
                                          e/d/1}
        \path[selected edge] (\source) -- node[weight] {$\weight$} (\dest);
\end{tikzpicture}
}
}
\subfigure[$k=9$]{
\scalebox{0.65}[0.65]{
\begin{tikzpicture}[scale=1.5, auto,swap]
    \foreach \pos/\name in {{(0,0)/a},{(2,0)/c}, 
                            {(0,-1)/d}, {(1,-1)/e}, {(2,-1)/f}}
        \node[vertexRegionA] (\name) at \pos {$\name$};
    \foreach \pos/\name in {{(1,0)/b}}
        \node[vertexRegionB] (\name) at \pos {$\name$};
    \foreach \source/ \dest /\weight  in {a/d/1, 
                                          c/f/8, 
                                          f/e/1,
                                          e/d/1}
        \path[selected edge] (\source) -- node[weight] {$\weight$} (\dest);
\end{tikzpicture}
}
}
\end{center}
\caption{Example of hierarchical image segmentations. In contrast to
  example in Fig.~\ref{pedro:nonhierarchical}, the contours are stable
  from a scale to another, providing  a hierarchy.}
\label{fig:our:hierarchical}
\end{figure}

Fig.~\ref{fig:our:hierarchical} illustrates the result of our method
on a pedagogical example. Starting from the graph of
Fig.~\ref{fig:our:hierarchical}(a), our method produces the
hierarchical observation scales depicted
in Fig.~\ref{fig:our:hierarchical}(b). 
As for the method of \cite{Felzenszwalb2004}, our algorithm only considers  the
edges of  the minimum spanning tree
(see~Fig.~\ref{fig:our:hierarchical}(c)). The whole hierarchy is
depicted as a dendrogram in Fig.~\ref{fig:our:hierarchical}(d), whereas
two levels of the hierarchy (at scales 2 and 9) are shown in
Fig.~\ref{fig:our:hierarchical}(e) and (f).

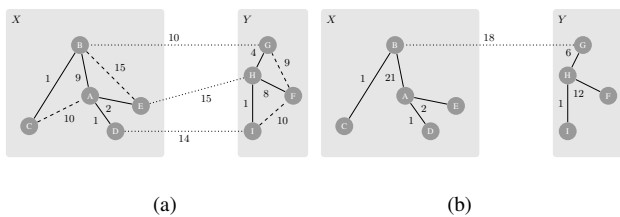
\begin{figure}
\begin{center}
\subfigure[]{
\hspace{-0.7cm} \scalebox{0.45}[0.45]{
\begin{tikzpicture}[scale=1.5, auto,swap  ]
    \node[vertexNAME] (C1) at (-1.45,1.5) {$\CompX$};
    \node[vertexNAME] (C2) at (3.1,1.5) {$\CompY$};

    \foreach \pos/\name /\pixel in {{(0,0)/a/1}, 
                                    {(-0.2,1.0)/b/12}, 
                                    {(-1.2,-0.6)/c/7},
                                    {(0.5,-0.7)/d/2}, 
                                    {(1,-0.2)/e/3}}
        \node[vertexG] (\name) at \pos {${\textsc{\name}}$};

    \foreach \pos/\name /\pixel in {{(4,0)/f/1}, 
                                    {(3.5,1)/g/12}, 
                                    {(3.2,0.4)/h/7},
                                    {(3.2,-0.7)/i/2}}
        \node[vertexG] (\name) at \pos {${\textsc{\name}}$};

    \foreach \pos/\name /\pixel in {{(-2,2)/XX/1}, 
                                    {(1.8,-1.5)/YY/12}, 
                                    {(3.0,2.0)/ZZ/7},
                                    {(4.2,-1.5)/WW/2}}
        \node[vertexNULL] (\name) at \pos {$$};
        
    \foreach \source/ \dest /\weight  in {b/c/1, b/a/9,
    																		  a/e/2,a/d/1}
        \path[edgeG] (\source) -- node[weight] {$\weight$} (\dest);

    \foreach \source/ \dest /\weight  in {g/h/4, h/f/8,
    																		  h/i/1}
        \path[edgeG] (\source) -- node[weight] {$\weight$} (\dest);

    \foreach \source/ \dest /\weight  in {e/h/15, g/b/10,
    																		  d/i/14}
        \path[edgeMG] (\source) -- node[weight] {$\weight$} (\dest);

    \foreach \source/ \dest /\weight  in {c/a/10, e/b/15,
    																		  f/g/9, i/f/10}
        \path[edgeEMG] (\source) -- node[weight] {$\weight$} (\dest);

  \begin{pgfonlayer}{background}
    \node [background2,fit=(XX) (YY)] {};
    \node [background2,fit=(WW) (ZZ)] {};

    \end{pgfonlayer}

\end{tikzpicture}
}
}
\hspace{-0.7cm}
\subfigure[]{
\scalebox{0.45}[0.45]{
\begin{tikzpicture}[scale=1.5, auto,swap  ]
    \node[vertexNAME] (C1) at (-1.45,1.5) {$\CompX$};
    \node[vertexNAME] (C2) at (3.1,1.5) {$\CompY$};

    \foreach \pos/\name /\pixel in {{(0,0)/a/1}, 
                                    {(-0.2,1.0)/b/12}, 
                                    {(-1.2,-0.6)/c/7},
                                    {(0.5,-0.7)/d/2}, 
                                    {(1,-0.2)/e/3}}
        \node[vertexG] (\name) at \pos {${\textsc{\name}}$};

    \foreach \pos/\name /\pixel in {{(4,0)/f/1}, 
                                    {(3.5,1)/g/12}, 
                                    {(3.2,0.4)/h/7},
                                    {(3.2,-0.7)/i/2}}
        \node[vertexG] (\name) at \pos {${\textsc{\name}}$};

    \foreach \pos/\name /\pixel in {{(-2,2)/XX/1}, 
                                    {(1.8,-1.5)/YY/12}, 
                                    {(3.0,2.0)/ZZ/7},
                                    {(4.2,-1.5)/WW/2}}
        \node[vertexNULL] (\name) at \pos {$$};
        
    \foreach \source/ \dest /\weight  in {b/c/1, b/a/21,
    																		  a/e/2,a/d/1}
        \path[edgeG] (\source) -- node[weight] {$\weight$} (\dest);

    \foreach \source/ \dest /\weight  in {g/h/6, h/f/12,
    																		  h/i/1}
        \path[edgeG] (\source) -- node[weight] {$\weight$} (\dest);

    \foreach \source/ \dest /\weight  in {g/b/18}
        \path[edgeMG] (\source) -- node[weight] {$\weight$} (\dest);


  \begin{pgfonlayer}{background}
    \node [background2,fit=(XX) (YY)] {};
    \node [background2,fit=(WW) (ZZ)] {};

    \end{pgfonlayer}

\end{tikzpicture}
}\vspace{-0.5cm} 
}

\end{center}

\caption{Example for computing the hierarchical scale for an edge-weighted graph. For this example, we suppose that all scales for the regions $\CompX$ and $\CompY$ are already computed, and we will calculate the hierarchical scale for the edge connectinge $B$ and $G$. 
}
\label{our:incrementalVolume}
\end{figure}

Let us illustrate the computation of a hierarchical obervation scale
on the graph of Fig.~\ref{our:incrementalVolume}(a). To this end, we
consider the iteration of the algorithm at which the edge $e$ linking
$B$ to $G$ is analyzed. At this step, the edges of the MST of weight
below $w(e)=10$ have been already processed. Therefore, the
hierarchical observation scale of these edges (depicted by continuous
lines in the figure) is already known as shown
in~\ref{our:incrementalVolume}(b). The regions~$X$ and~$Y$ obtained at
steps (i) and (ii) are set to $\{A,B,C,D,E\}$ and~$\{F,G,H,I\}$
respectively. Then, in order to find the value~$\Lowlevel{e}$ at step
(iii), the partitions $\PARTITIONSETGW[i]{\G}{\W}$ for $i
=\{2,7,13,18\}$ must be considered. We have:
$\PARTITIONSETGW[2]{\G}{\SCALESETSYMBOL}=\{\{B,C\},\{A,D\},\{E\},\{G\},\{F\},\{I,H\}\}$,
$\PARTITIONSETGW[7]{\G}{\SCALESETSYMBOL}=\{\{B,C\},\\
\{A,D,E\},\{F\},\{G,H,I\}\}$,
$\PARTITIONSETGW[13]{\G}{\SCALESETSYMBOL}=\{\{B,C\},\{A,D,E\},\{F,G,\\
H,I\}\}$ and
$\PARTITIONSETGW[18]{\G}{\SCALESETSYMBOL}=\{\{B,C\},\{A,D,E\},\{F,G,H,I\}\}$.
By the application of steps (1-5), the value~$\NRINT[\CompY]{\CompX}$
is found to be $18$ since $18$ is the first value below the observation
scale of the region containing $B$ relatively to~$Y$.
The same process is made for $\NRINT[\CompX]{\CompY}$, but the regions
are $\{G\}$, $\{G,H,I\}$ and $\{G,H,I,F\}$. Moreover, the observation
scale is $12$ since $12$ is the first value below the observation
scale of the region containing $G$ relatively to~$X$.
Finally, the observation scale of $\CompX$ and $\CompY$ is
$18$. 

To efficiently implement our method, we use some data structures
similar to the ones proposed in \cite{Cousty2011}; in particular, the
management of the collection of partitions are due to Tarjan's union
find. Furthermore, we made some algorithmic optimizations to speed up
the computations of the observation scales. In order to illustrate an
example of computation time, we implemented all our algorithm in C++
on a standard single CPU computer under windows Vista, we run it in a
Intel Core 2 Duo, 4GB. For the image illustrated in
Fig.~\ref{pedro:real:nonhierarchical}(a) (with size 321x481), the
hierarchy is computed in $2.7$ seconds, and the method proposed in
\cite{Felzenszwalb2004} spent $1.3$ seconds .


\section{Experimental results} \label{sec:experiment}

A major difficulty of experiments is the design of an adequate
edge-cost, well adapted to the content to be segmented.  A practical
solution is to use some dissimilarity functions, and many different
approaches are used in the litterature. In this work, the underlying
graph is the one induced by the 4-adjacency relation, and the edges
are weighted by a simple color gradient computed by an Euclidean
distance in the RGB space.

\begin{figure}
\begin{center} 	\renewcommand{\tabcolsep}{0.1cm}
\subfigure[]{\includegraphics[width=0.13\linewidth]{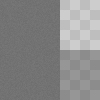}}
\subfigure[]{\includegraphics[width=0.13\linewidth]{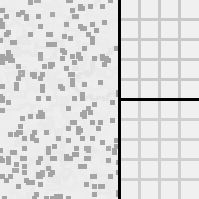}}
\subfigure[41]{\includegraphics[width=0.13\linewidth]{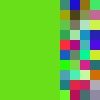}}
\subfigure[22]{\includegraphics[width=0.13\linewidth]{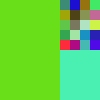}}
\subfigure[3]{\includegraphics[width=0.13\linewidth]{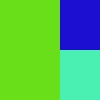}}
\subfigure[2]{\includegraphics[width=0.13\linewidth]{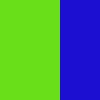}} 
\subfigure[1]{\includegraphics[width=0.13\linewidth]{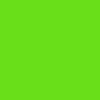}}



\end{center}
\caption{\label{fig:results:sintetico}An example of a hierarchical
  image segmentations of a synthetic image containinng three
  perceptually big regions.\COMMENT{, however, depending on the scale,
    two of them can segmented into smallest regions.} The saliency map
  of the image (a) is showned in (b). \COMMENT{The images are obtained
    at scales (c) 1000, (d) 2000, (e) 5000, (f) 140000 and (g)
    242000.} The number of regions of the segmented images is written
  under each figure.}
\end{figure}

In Fig.~\ref{fig:results:sintetico}, we present some results on an
artifical image containing three perceptually big regions. On this
example, one can easily verify the hierarchical property of our method
by looking at the segmentations at scales {\em resp.} 1000, 2000,
5000, 140000 and 224000 ({\em resp.}
Fig.~\ref{fig:results:sintetico}(c), (d), (e), (f) and (g)). Since the
resulting segmentations are nested, the whole hierarchy can be
presented in a saliency map (see Fig. \ref{fig:results:sintetico}(b)).

\COMMENT{The hierarchy of Fig.~\ref{fig:results:sintetico}(a) has 45
  scales (see the saliency map Fig.~\ref{fig:results:sintetico}(b)).}

\begin{figure}
\begin{center} 	\renewcommand{\tabcolsep}{0.1cm}
\begin{tabular}{cccc} 
\includegraphics[width=0.23\linewidth]{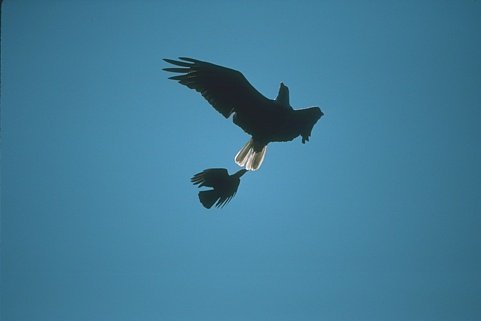} &
\includegraphics[width=0.23\linewidth]{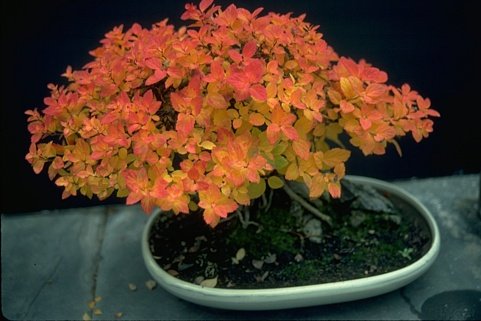} &
\includegraphics[width=0.23\linewidth]{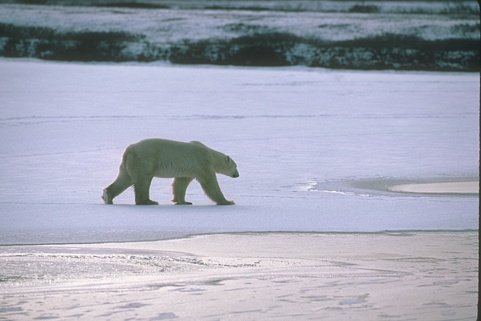} &
\includegraphics[width=0.23\linewidth]{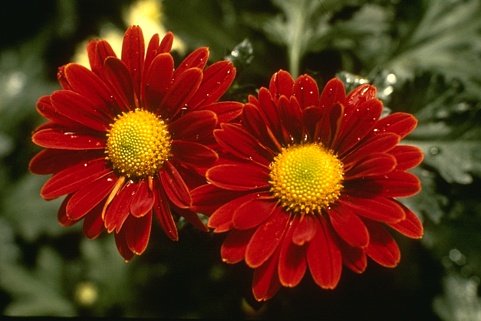} \\ 
{\includegraphics[width=0.23\linewidth]{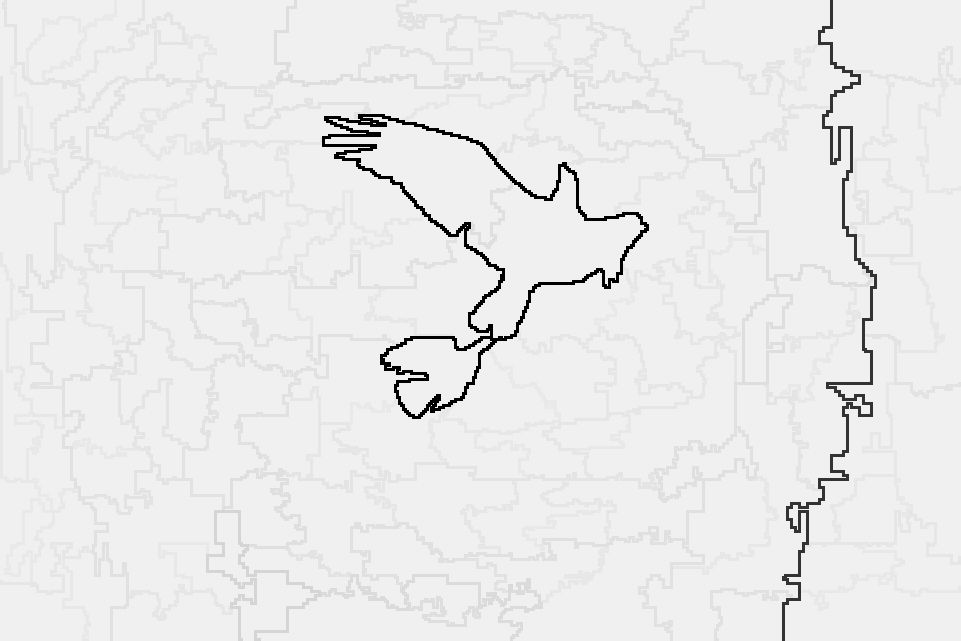}}&
{\includegraphics[width=0.23\linewidth]{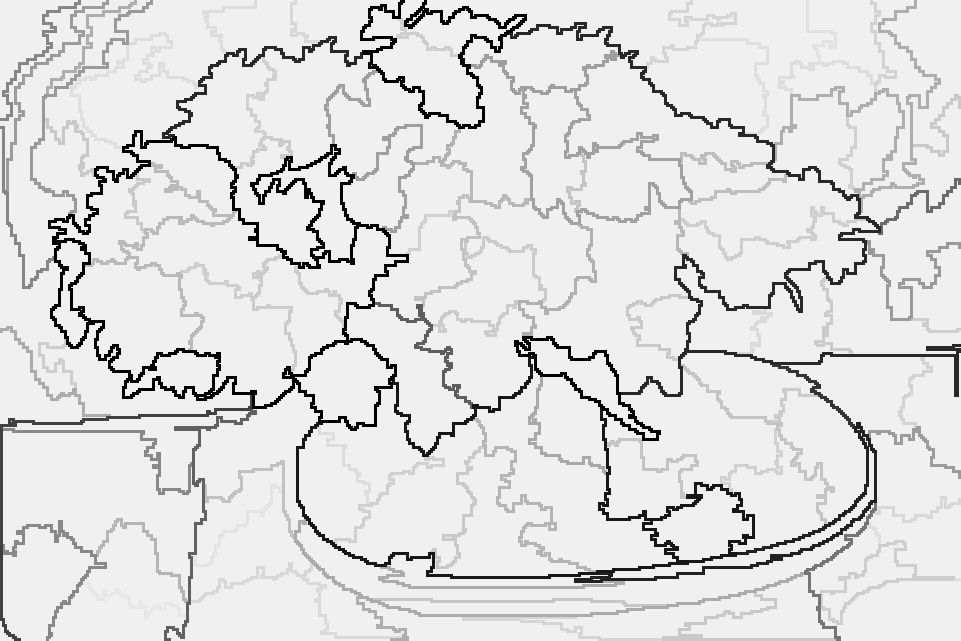}}& 
{\includegraphics[width=0.23\linewidth]{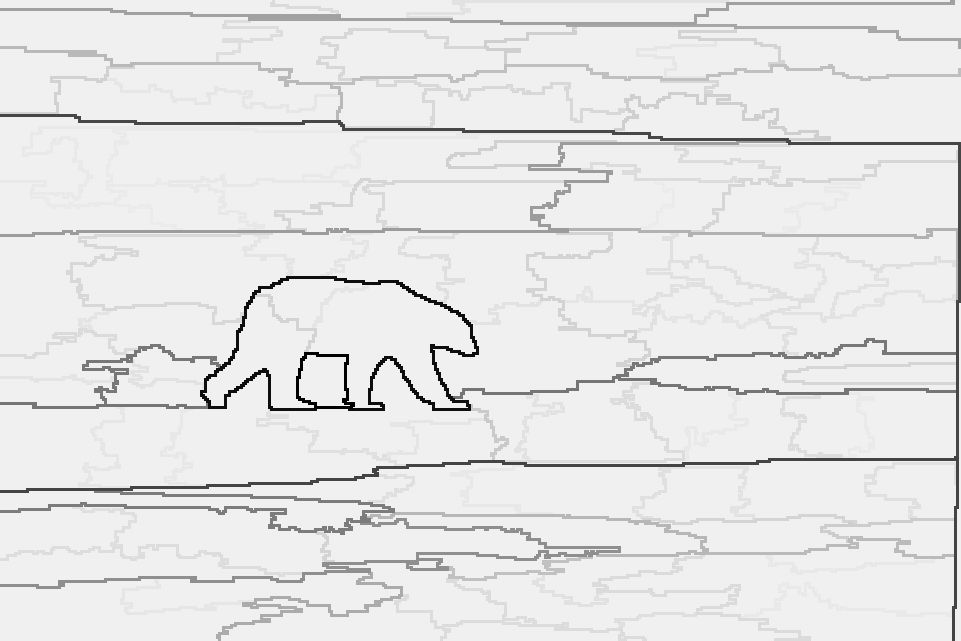}}&
{\includegraphics[width=0.23\linewidth]{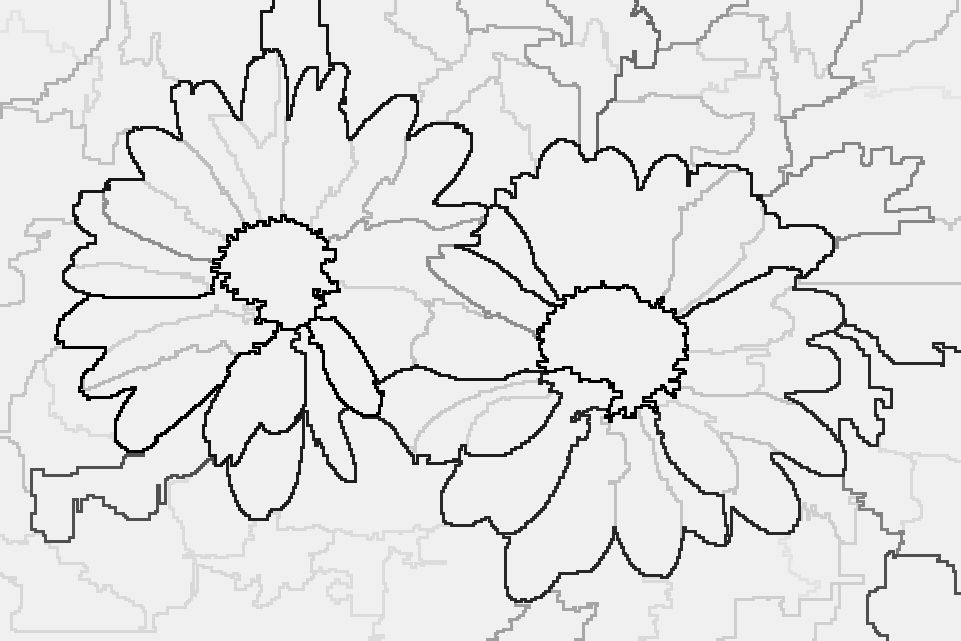}} \\
{\includegraphics[width=0.22\linewidth]{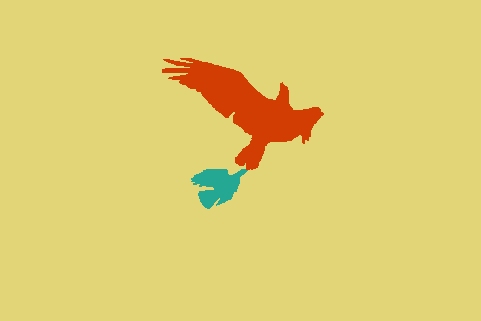}}&
{\includegraphics[width=0.22\linewidth]{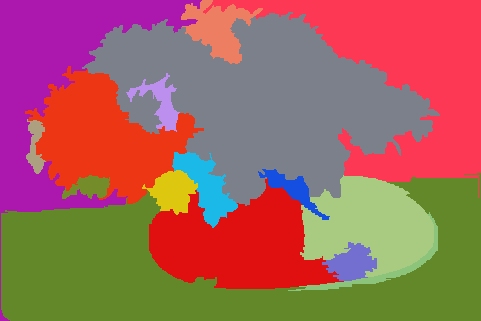}}&
{\includegraphics[width=0.22\linewidth]{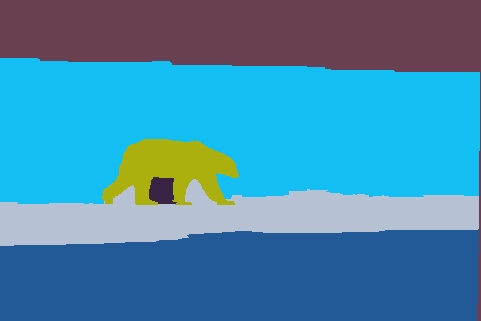}}&
{\includegraphics[width=0.22\linewidth]{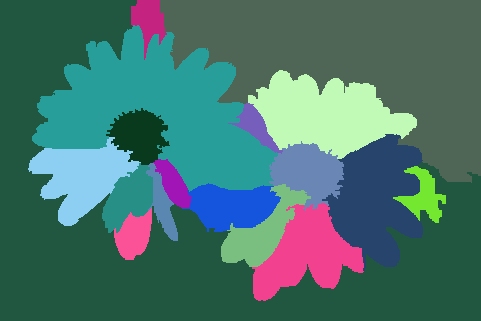}}\\
(a) & (b) & (c) & (d) \\ 
\end{tabular}
\end{center}
\caption{\label{fig:results:salliance}Top row: some images of the
  Berkeley database~\cite{MartinFTM01}.  Middle row: saliency maps of
  these images. The numbers of scales of these hierarchies are (a) 240,
  (b) 429, (c) 405 and (d) 443.
  Bottom row: according to our subjective judgment, the best
  segmentations extracted from the hierarchies. The numbers of regions
  are (a) 3, (b) 16, (c) 6 and (d) 18. }
\end{figure}

Fig.~\ref{fig:results:salliance} illustrates the performance of our
method on some images of the Berkeley's database~\cite{MartinFTM01}.
Note that, as in \cite{Felzenszwalb2004}, an area filtering is applied
to eliminate small regions (smaller than 500 pixels).

\begin{figure}
\begin{center} 	\renewcommand{\tabcolsep}{0.1cm}
\subfigure[16]{
\includegraphics[width=0.23\linewidth]{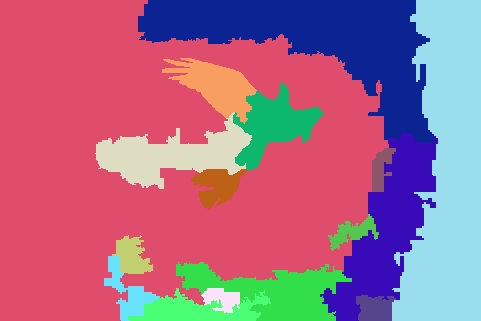}
\includegraphics[width=0.23\linewidth]{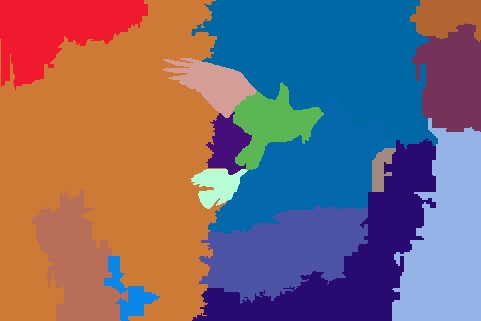}
}
\subfigure[52]{
\includegraphics[width=0.23\linewidth]{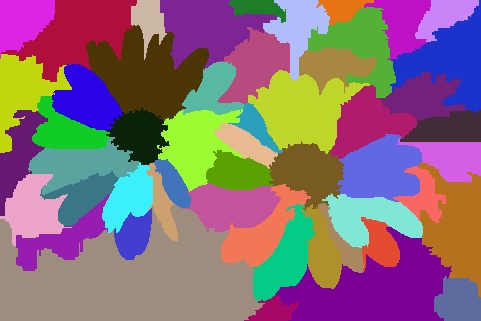}
\includegraphics[width=0.23\linewidth]{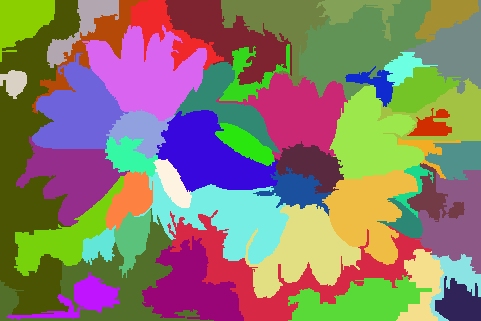}
}
\subfigure[26]{
\includegraphics[width=0.23\linewidth]{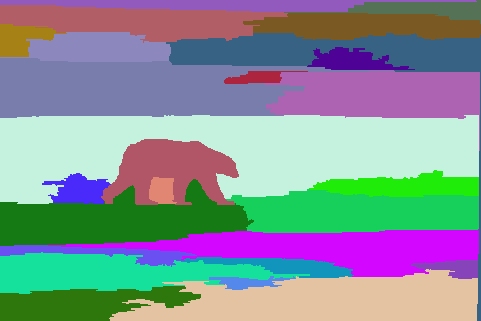}
\includegraphics[width=0.23\linewidth]{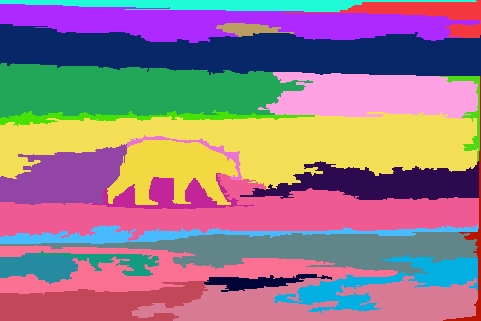}
}
\subfigure[18]{
\includegraphics[width=0.23\linewidth]{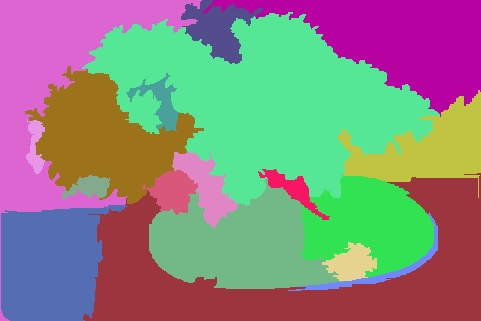}
\includegraphics[width=0.23\linewidth]{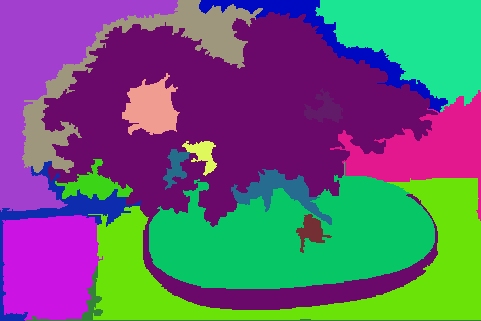}}
\end{center}
\caption{Comparison between \cite{Felzenszwalb2004} and our
  approach. For each pair of images, the right image shows the best
  result (according to our judgment and our experiments) from
  \cite{Felzenszwalb2004} and the left image shows a segmentation
  extracted from our hierarchical result, with the same number of
  regions.
  \label{fig:results:regions}}
\end{figure}

\begin{figure}
\begin{center} 	\renewcommand{\tabcolsep}{0.1cm}
\subfigure[]{
\includegraphics[width=0.23\linewidth]{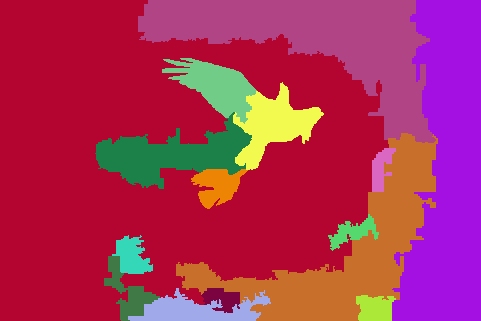}
\includegraphics[width=0.23\linewidth]{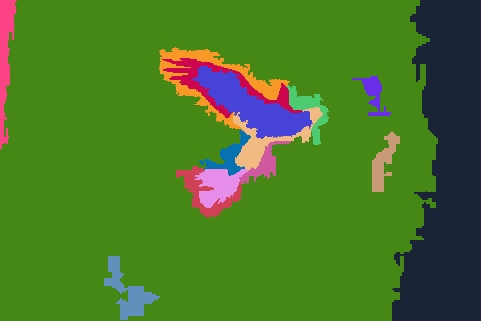}
}
\subfigure[]{
\includegraphics[width=0.23\linewidth]{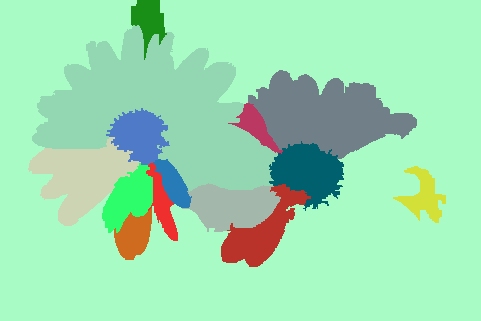}
\includegraphics[width=0.23\linewidth]{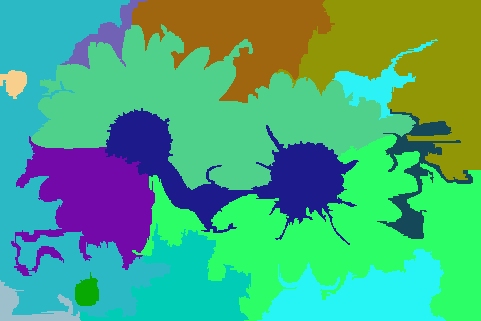}
}
\subfigure[]{
\includegraphics[width=0.23\linewidth]{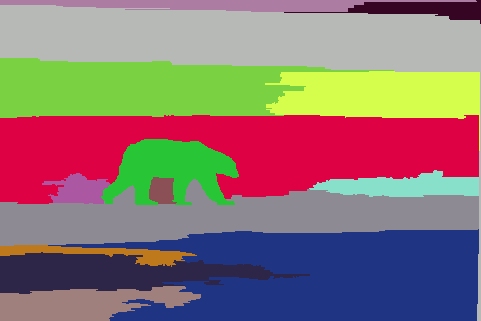}
\includegraphics[width=0.23\linewidth]{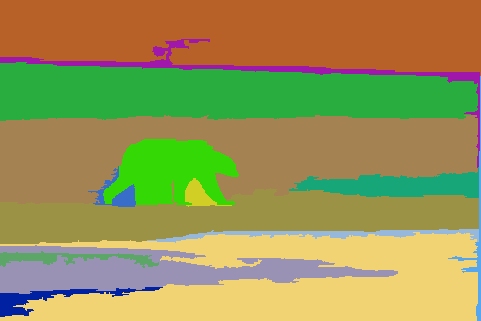}
}
\subfigure[]{
\includegraphics[width=0.23\linewidth]{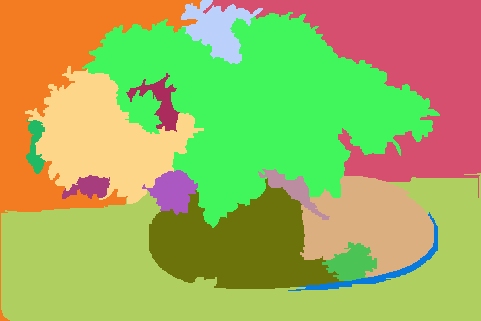}
\includegraphics[width=0.23\linewidth]{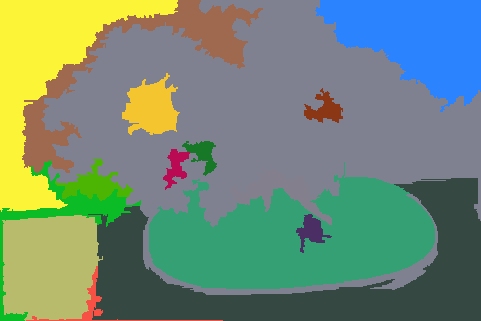}}
\end{center}
\vspace{-0.5cm}
\caption{Examples of image segmentation where the number of regions
  has been set to 15. For each pair of images, the left one shows a
  segmentation extracted from our hierarchy, with the desired number
  of regions; and the right one shows the result obtained with
  \cite{Felzenszwalb2004} by varying the parameter $k$ until the
  desired number of regions is found. \label{fig:results:x:regions}}
\end{figure}

\begin{figure}
\begin{center} 	\renewcommand{\tabcolsep}{0.1cm}
\begin{tabular}{ccc} 
\includegraphics[width=0.30\linewidth]{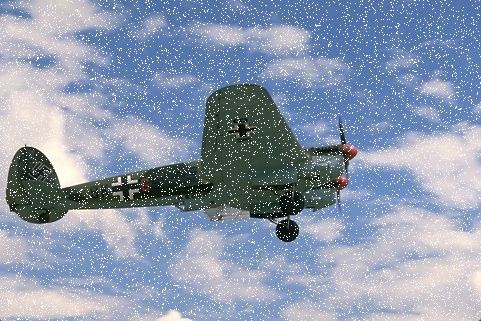} &
\includegraphics[width=0.30\linewidth]{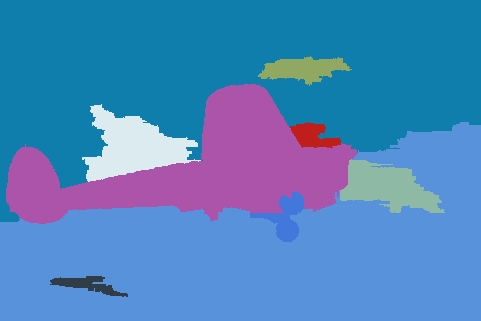}&
\includegraphics[width=0.30\linewidth]{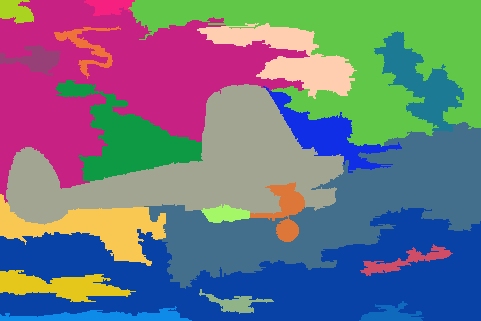}\\
\includegraphics[width=0.30\linewidth]{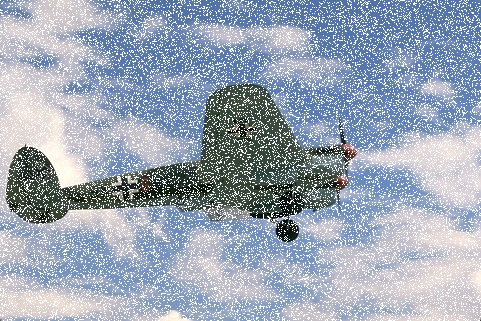} &
\includegraphics[width=0.30\linewidth]{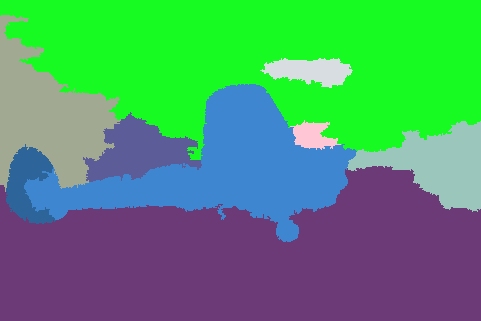}&
\includegraphics[width=0.30\linewidth]{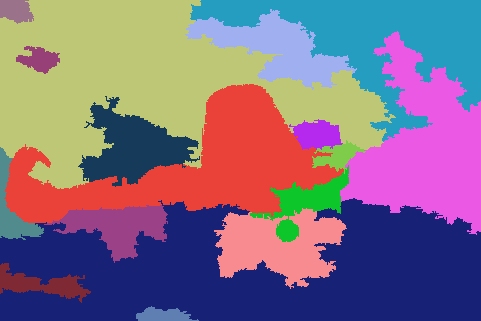}\\
\includegraphics[width=0.30\linewidth]{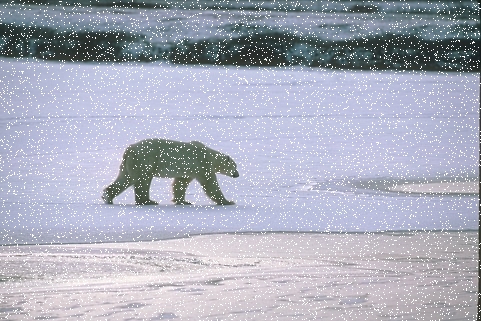} &
\includegraphics[width=0.30\linewidth]{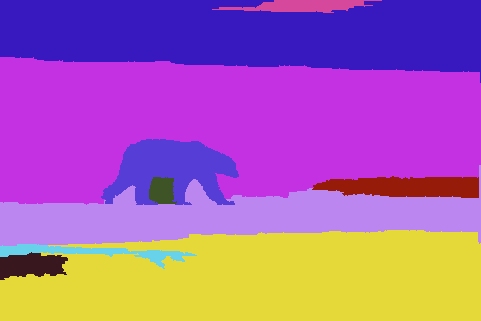}&
\includegraphics[width=0.30\linewidth]{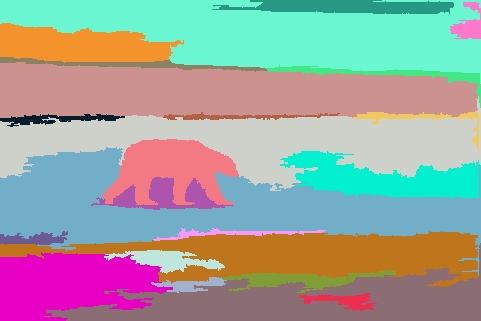}\\
\includegraphics[width=0.30\linewidth]{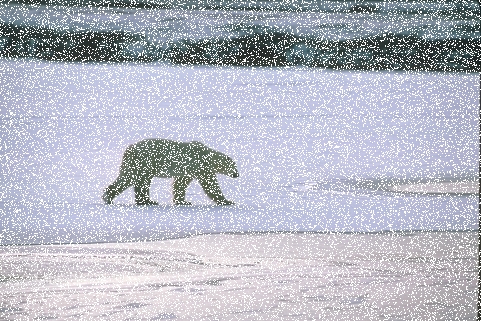} &
\includegraphics[width=0.30\linewidth]{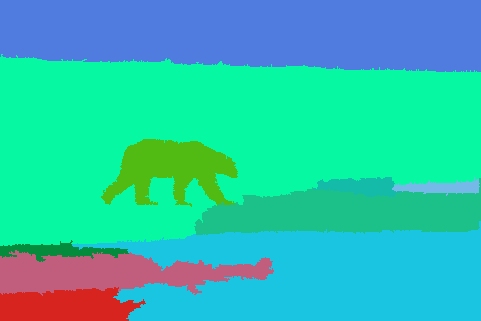}&
\includegraphics[width=0.30\linewidth]{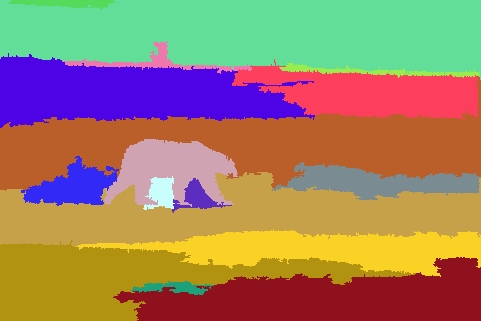}\\
(a) & (b) & (c) \\
\end{tabular}
\end{center} 
\caption{Examples of segmentations for images corrupted by a random
  salt noise. The corrupted images (at different levels - $70\%$ and
  $90\%$) are shown on the first column. The results of our method
  and \cite{Felzenszwalb2004} are illustrated in the second and third
  columns, respectively. \label{fig:results:noise} }
\end{figure}

Comparison of the results of our algorithm with the ones of
\cite{Felzenszwalb2004} are difficult, since the tuning of the
parameters of \cite{Felzenszwalb2004} is critical and since we produce
a whole hierarchy of segmentations. We made three experiments. First,
we try to set the correct parameter for \cite{Felzenszwalb2004}, {\em
  i.e.} the parameter that produces the best (subjective) visual
result (Fig.~\ref{fig:results:regions}). We can compare this result
with on the one hand, the ``best'' segmentation extracted from our
hierarchy in Fig.~\ref{fig:results:salliance}, and on the other hand,
with a segmentation from our hierarchy containing the same number of
regions as \cite{Felzenszwalb2004} (Fig.~\ref{fig:results:regions}).
In a second experiments, we fixed the number of regions to 15 for all
images, and tune the parameter of \cite{Felzenszwalb2004} to obtain
this number of regions. We can compare these segmentations with our
own results on Fig.~\ref{fig:results:x:regions}. The last experiments
is designed to assess the robustness to random impulse noise, see
Fig.~\ref{fig:results:noise}.

\section{Conclusions} \label{sec:conclusions}
This paper proposes an efficient hierarchical segmentation method
based on the observation scales of \cite{Felzenszwalb2004}. In
contrast to \cite{Felzenszwalb2004}, our method produces the complete
set of the segmentations at every scales, and satisfies both the
causality and location principle defined by \cite{Guigues2006}. An
important practical consequence of these properties is to ease the
selection of a scale level adapted to a particular task. We visually
assessed our method on some real images by comparing our segmentations
to those of \cite{Felzenszwalb2004}. Even if more (quantitative) tests
(such as the ones proposed by \cite{Arbelaez2011}) are needed for
drawing definitive conclusions, the produced segmentations are
promizing, in particular w.r.t. robustness. As future work, we will
investigate using more information into the definition of observation
scale as well as learning which information is pertinent for a given
practical task.


\bibliographystyle{IEEEbib}
\bibliography{mst}

\end{document}